\documentclass[letterpaper, 10 pt, conference]{ieeeconf}  

\IEEEoverridecommandlockouts                              

\usepackage{amsmath}
\usepackage{graphicx}
\usepackage{amssymb}
\usepackage{booktabs}
\usepackage{multirow}
\usepackage{multicol}
\usepackage{float}
\usepackage{subfigure}
\usepackage{kotex}
\usepackage{tabularx}
\usepackage{array}
\usepackage[flushleft]{threeparttable}
\usepackage[ruled,vlined]{algorithm2e}
\usepackage{bm}
\usepackage{breqn}
\usepackage{footnote}
\usepackage{cite}
\makeatletter
\let\NAT@parse\undefined
\makeatother
\usepackage{hyperref}
\usepackage{xcolor}

\begin{document}
\definecolor{edit}{RGB}{0,0,0}

%
\title{UV-SLAM: Unconstrained Line-based SLAM Using Vanishing Points for Structural Mapping}
%
%
%

\author{Hyunjun Lim$^{1}, $~\IEEEmembership{Student~Member, IEEE}, Jinwoo Jeon$^{1}, $~\IEEEmembership{Student~Member, IEEE},\\ and Hyun Myung$^{2*}, $~\IEEEmembership{Senior~Member, IEEE}%
\thanks{$^{1}$H. Lim and J. Jeon are with School of Electrical Engineering, Korea Advanced Institute of Science and Technology~(KAIST), Daejeon, Republic of Korea
{\tt\small \{tp02134, zinuok\}@kaist.ac.kr}}%
\thanks{$^{2}$H. Myung is with School of Electrical Engineering, KI-AI, and KI-R, KAIST, Daejeon, Republic of Korea
{\tt\small hmyung@kaist.ac.kr}}%
\thanks{This work has been supported by the Unmanned Swarm CPS Research Laboratory program of Defense Acquisition Program Administration and Agency for Defense Development. (UD190029ED)}
}

%
%

\markboth{IEEE Robotics and Automation Letters. Preprint Version. Accepted Month, Year}
{FirstAuthorSurname \MakeLowercase{\textit{et al.}}: ShortTitle} 
%



\maketitle

\begin{abstract}
In feature-based simultaneous localization and mapping (SLAM), line features complement the sparsity of point features, making it possible to map the surrounding environment structure. Existing approaches utilizing line features have \textcolor{edit}{primarily} employed a measurement model that uses \textcolor{edit}{line} re-projection. However, the direction vectors used in the 3D line mapping process cannot be corrected because the line measurement model employs only the lines' normal vectors in the Pl\"{u}cker coordinate. As a result, problems like degeneracy that occur during the 3D line mapping process cannot be solved. To tackle the problem, this paper presents a UV-SLAM, which is an unconstrained line-based SLAM using vanishing points for structural mapping. This paper focuses on using structural regularities without any constraints, such as the Manhattan world assumption. For this, we use the vanishing points that can be obtained from the line features. The difference between the vanishing point observation calculated through line features in the image and the vanishing point estimation calculated through the direction vector is defined as a residual and added to the cost function of optimization-based SLAM. \textcolor{edit}{Furthermore}, through Fisher information matrix rank analysis, we prove that vanishing point measurements guarantee a unique mapping solution. Finally, we demonstrate that the localization accuracy and mapping quality are improved compared to the state-of-the-art algorithms using public datasets.
\end{abstract}


%
\IEEEpeerreviewmaketitle

\section{Introduction}
The feature-based visual simultaneous localization and mapping (SLAM) has \textcolor{edit}{been primarily} developed based on point features because a point is the smallest unit that can express the characteristics of an image and has an advantage in low computation environment. In addition, point features have been used for localization because they are easy to track. However, point features have several drawbacks\cite{he2018pl}. First, \textcolor{edit}{point features are} not robust in low-texture environments such as hallways. \textcolor{edit}{Moreover}, \textcolor{edit}{they are} weak against illumination change. Finally, point features are sparse, \textcolor{edit}{making it} difficult to visualize the surrounding environment with a 3D map.

\begin{figure}[t]
    \centering
    \subfigure[]{\label{fig:before}\includegraphics[width=\linewidth]{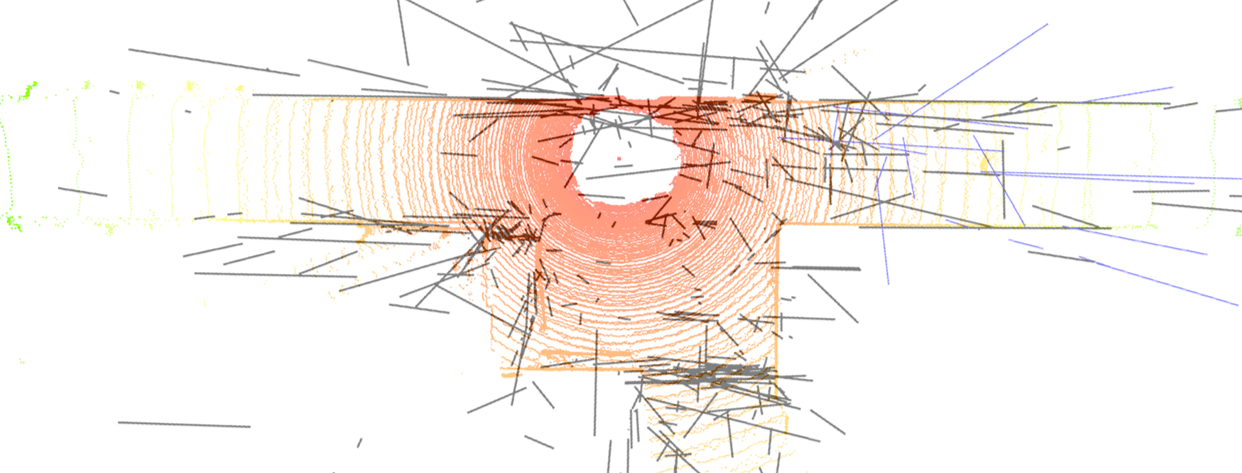}}
    \subfigure[]{\label{fig:after}\includegraphics[width=\linewidth]{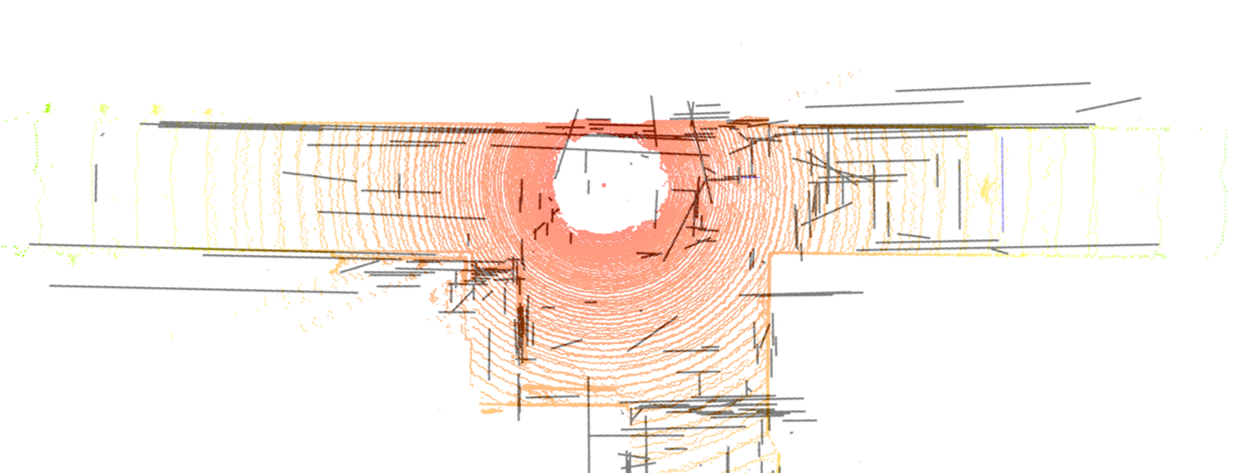}}
    \caption{Top views of 3D line mapping results with LiDAR point clouds overlaid in real corridor environment \subref{fig:before} ALVIO\cite{jung2020alvio,lim2021avoiding}. \subref{fig:after} UV-SLAM.}
    \label{fig:introduction}
    \vspace{-5mm}
\end{figure}

To supplement the point features, line-based methods have been proposed. Line features can additionally be used in environments with low textures, such as corridors. In addition, \textcolor{edit}{because} the line consists of several points, there is a high \textcolor{edit}{probability} that the characteristics will be maintained even when an illumination change occurs\cite{kottas2013efficient,kong2015tightly}. Finally, because line features have structural regularities, the surrounding environment can be easily identified through 3D mapping\cite{zhang2015building}.

With the above advantages, many studies have been conducted to apply line features to visual SLAM. First, a method using the Pl\"{u}cker coordinate and the orthonormal representation for representing 3D lines was proposed\cite{bartoli2005structure} to express a 3D line with a higher \textcolor{edit}{degree of freedom} (DoF) than a 3D point. Based on this, line measurement model was defined in a similar way to the point measurement model. It re-projects a 3D line and calculates the difference from the new observed line. Most line-based algorithms adopt similar methods on existing point-based methods. Filtering-based approaches\cite{zheng2018trifo,8967905} were developed from MSCKF\cite{mourikis2007multi}. Some optimization-based methods\cite{pumarola2017pl,zuo2017robust,lee2019elaborate} exploited ORB-SLAM\cite{mur2015orb} and other approaches\cite{he2018pl,fu2020pl,jung2020alvio,lim2021avoiding} used VINS-Mono\cite{qin2018vins}.


However, the above algorithms applying only line measurement model does not solve the problems in the 3D mapping of lines. Among the difficulties, a degeneracy problem occurs in the 3D mapping of line features. The degeneracy refers to a phenomenon in which 3D features cannot be uniquely determined through the triangulation of features\cite{hartley2003multiple}. When the observed 2D point is close to the epipole, a point feature cannot be determined as a single 3D point because the 3D point exists on the baseline. Similarly, when the observed 2D line passes close to the epipole, a line feature cannot be determined as a single 3D line because the 3D line exists on the epipolar plane. The mapping of line features is inaccurate because degeneracy occurs more frequently than point features \textcolor{edit}{do}. However, because the line measurement model used in the existing algorithm employs only each line's normal vector in the Pl\"{u}cker coordinate, their direction vectors cannot be corrected. Fig.~\ref{fig:introduction} shows the comparison of top views of 3D line mapping results between ALVIO\cite{jung2020alvio,lim2021avoiding} and the proposed algorithm with the LiDAR point clouds overlaid in a real corridor environment. The ALVIO has poor mapping result despite the lines having structural regularities as shown in Fig.~\ref{fig:before}. 

In this paper, we propose a UV-SLAM, an algorithm that solves the above problems using the lines' structural regularities as shown in Fig.~\ref{fig:after}. The main contributions of this paper are as follows:
\begin{itemize}
    \item To the best of our knowledge, the proposed UV-SLAM is the first optimization-based monocular SLAM using vanishing point measurements for structural mapping without any restriction such as camera motion and environment. In particular, our algorithm does not use the Manhattan world assumption in the process of extracting the vanishing points and using \textcolor{edit}{them} as measurement\textcolor{edit}{s}.
    \item We define a novel residual term and Jacobian \textcolor{edit}{of the} vanishing point measurements based on the most common methods of expressing 3D line features\textcolor{edit}{:} the Pl\"{u}cker coordinate and the orthonormal representation. 
    \item We prove that the proposed method guarantees the observability of 3D lines through Fisher information matrix (FIM) rank analysis. Through this, problems that occur in the 3D mapping of line features are proven to be solved.
\end{itemize}

The rest of this paper is organized as follows. Section~\ref{sec:related work} gives a review of related works. Section~\ref{sec:proposed method} explains the proposed method in depth. \textcolor{black}{Section~\ref{sec:FIM rank analysis} analyzes the Fisher information matrix rank to prove the validity of the proposed method.} Section~\ref{sec:experimental results} provides the experimental results. Finally, Section~\ref{sec:conclusion} concludes by summarizing our contributions and discussing future work.

\section{Related works}\label{sec:related work}
\subsection{Degeneracy of Line Features}
Some works looked into the degeneracy for line features\cite{sugiura20153d,zhou2018slam}. In addition, Yang \textit{et al.} investigated degenerate motions using two distinct line triangulation methods\cite{8967905}. Subsequently, they analyzed various features' observability and degenerate camera motions in the inertial measurement unit (IMU) aided navigation system\cite{yang2019observability}. However, when line degeneracy occurs in these investigations, all of the degenerate lines have been eliminated. As a result, there is a limitation in that information loss occurs due to the removed lines.

To handle degenerate lines, Ok \textit{et al.} reconstructed 3D line segments using imaginary points when the lines are close to the epipolar line\cite{ok2012accurate}. However, this approach can be used only when degenerate lines intersect with other lines, and it restricts applicability. Our previous work solved the degeneracy by using structural constraints in parallel conditions, investigating the fact that degenerate lines frequently occur in pure translation motions\cite{lim2021avoiding}. However, this method has a limitation in that it can be used only in pure translational motions. \textcolor{black}{Therefore, in order to improve the quality of line mapping, a method that can be used independent of camera motion is required.}

\textcolor{black}{\subsection{Line-based SLAM with Manhattan or Atlanta World Assumption}}
In \cite{kim2018low,kim2018indoor}, rotation matrix was estimated using the Manhattan world assumption. Based on this, decoupled methods have been proposed to estimate the translation after calculating the rotation through the vanishing points\cite{li2020structure,li2020rgb,yunus2021manhattanslam}. \textcolor{black}{In addition, some approaches applied line features to SLAM by using the Manhattan or Atlanta world assumption \cite{zhou2015structslam,zou2019structvio}.} These methods use a novel 2-DoF line representation to exploit lines with structural regularities only. \textcolor{black}{Furthermore, there is a study using 2-DoF line representation to classify structural lines and non-structural lines\cite{xu2021leveraging}.} However, as these approaches use structural lines with dominant direction only, they are practical only in an indoor environment where the assumptions are mostly correct. \textcolor{black}{Therefore, there is a need for a novel algorithm that is not restricted by assumptions.}\\

\subsection{Vanishing Point Measurements}
Some approaches use vanishing point measurements without assumptions. In \cite{lee2021plf}, parallel lines were clustered based on vanishing points. Then, residuals were constructed using the conditions that parallel lines should be in one plane and their cross product should be zero. However, accurate mapping results could not be obtained when the initial estimation was inaccurate as degeneracy occurred. 

\textcolor{black}{Moreover, there is a paper using vanishing points as an observation model. In \cite{ma2019line}, the residuals are defined to apply unbounded vanishing point measurements to line-based SLAM. Unfortunately, it does not provide a proof that vanishing point measurement improves the localization accuracy and line mapping results of line-based SLAM.}

\section{Proposed method}\label{sec:proposed method}
\begin{figure*}[t]
    \centering
    \includegraphics[width=0.75\linewidth]{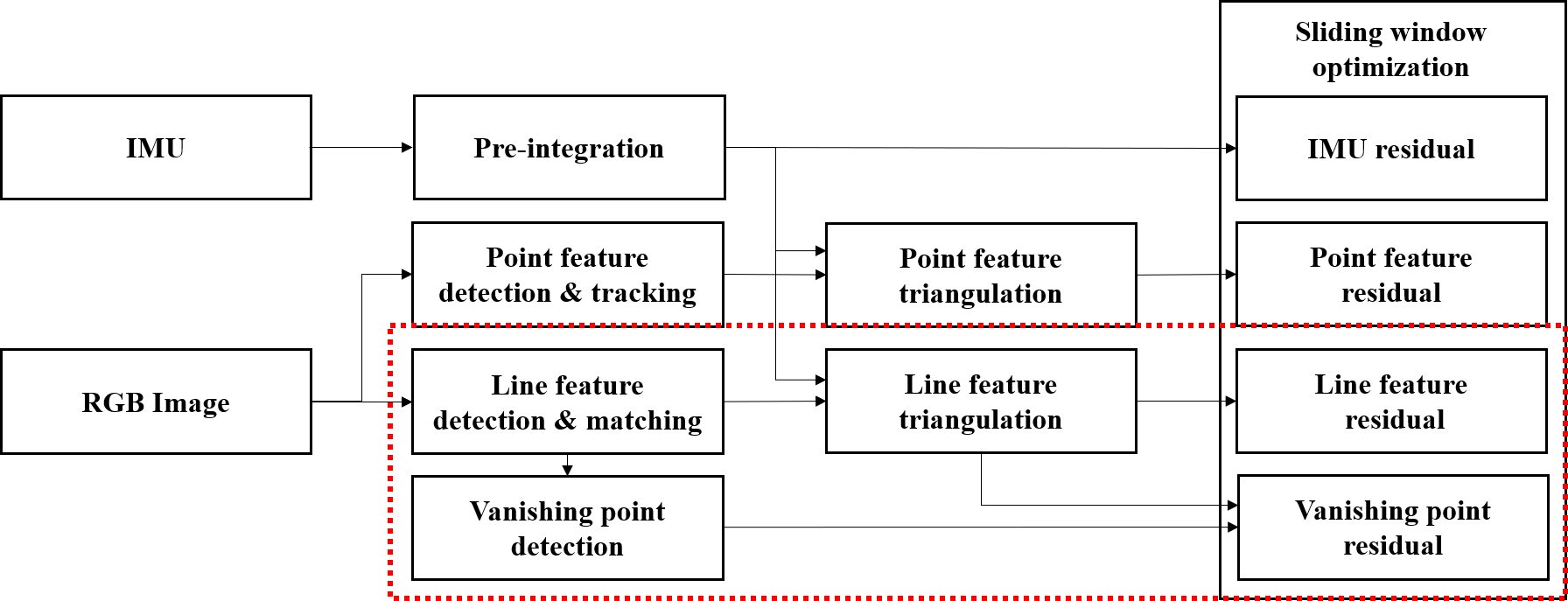}
    \caption{Block diagram illustrating the framework of UV-SLAM. The dashed box represents the newly added blocks in this paper. When an RGB image is received, detection and matching of line features are carried out. \textcolor{edit}{Afterward}, the vanishing points are detected\textcolor{edit}{,} and each line is clustered. After creating 3D lines through the triangulation process, the residuals of both the lines and the vanishing points are defined. Finally, localization and mapping results can be obtained through sliding window optimization.}
    \vspace{-5mm}
    \label{fig:system_overview}
\end{figure*}

\subsection{Framework of Algorithm}
The overall framework of the UV-SLAM is shown in Fig.~{\ref{fig:system_overview}}. The proposed method is based on VINS-Mono\cite{qin2018vins}, and the way IMU and point measurements are used is similar to it. The point features are extracted from Shi-Tomasi\cite{shi1994good} and are tracked by KLT\cite{tomasi1991detection}. In addition, the IMU measurement model is defined by the pre-integration method\cite{forster2016manifold}. Finally, our optimization-based method employs two-way marginalization with Schur complement\cite{sibley2010sliding}.

\textcolor{edit}{To} add line features to the monocular visual-inertial odometry (VIO) system, the line features are extracted from line segment detector (LSD)\cite{von2008lsd} and \textcolor{edit}{are} tracked by line binary descriptor (LBD)\cite{zhang2013efficient}. Whereas 3D points can be intuitively expressed in $(x,y,z)$, 3D lines require a complicated way to express themselves. Therefore, the proposed algorithm employs a Pl\"{u}cker coordinate and a orthonormal representation used in \cite{bartoli2005structure}. \textcolor{edit}{The} Pl\"{u}cker coordinate is an intuitive way to represent 3D lines, and a 3D line \textcolor{edit}{is represented} as follows: 
\begin{equation}
\mathbf{L}({\mathbf{n}}, {\mathbf{d}})^{\top}\in\mathbb{R}^6,
\end{equation} 
where $\mathbf{n}$ and $\mathbf{d}$ represent normal and direction vectors, respectively. The Pl\"{u}cker coordinate is used in the triangulation and re-projection process. Whereas 3D lines are actually 4-DoF, the lines in the Pl\"{u}cker coordinate are 6-DoF. Therefore, over-parameterization problem occurs in the optimization process of VIO or visual SLAM. To solve this, the orthonormal representation is employed, which is a 4-DoF representation of lines. It is used in the optimization process and can be expressed as follows:
\begin{equation}
\mathbf{o}=[\boldsymbol{\psi}, \phi],    
\end{equation}
where $\boldsymbol{\psi}$ is the 3D line's rotation matrix in Euler angles with respect to the camera coordinate system, and $\phi$ is the parameter representing the minimal distance from the camera center to the line. The conversion between the Pl\"{u}cker coordinate and the orthonormal representation is given in \cite{bartoli2005structure}.

In addition, the UV-SLAM can determine whether the extracted lines have structural regularities or not. For vanishing point detection, the proposed algorithm use J-linkage\cite{toldo2008robust}, which can find multiple instances in the presence of noise and outliers. The overall process is as follows: First, vanishing point hypotheses are created through random sampling for all lines extracted from the image. Subsequently, after merging similar ones through comparison between the hypotheses, vanishing points are calculated. Because the J-linkage can find all possible vanishing points through the hypotheses, it can find more vanishing points than other algorithms with the Manhattan world assumption.

\subsection{State Definition}
In this paper, $(\cdot)^w$, $(\cdot)^c$, and $(\cdot)^b$ represent the world coordinate, camera coordinate, and body coordinate, respectively. In addition, $(\cdot)^w_b$ \textcolor{edit}{reflects} the coordinate transformations of a rotation matrix, quaternion, or translation from the body coordinate to the world coordinate. The state vector used in our system is as follows:
\begin{equation}
\begin{aligned}
\mathcal{X} = [&\mathbf{x}_{0}, \mathbf{x}_{1}, \cdots ,\mathbf{x}_{I-1},\\
&\lambda_{0}, \lambda_{1}, \cdots, \lambda_{J-1},\\ 
&\mathbf{o}_{0}, \mathbf{o}_{1}, \cdots, \mathbf{o}_{K-1}],\\
\mathbf{x}_i = [&\mathbf{p}^{w}_{b_i}, \mathbf{q}^{w}_{b_i}, \mathbf{v}^{w}_{b_i}, \mathbf{b}_{a}, \mathbf{b}_{g}],~i\in[0,I-1],\\
\mathbf{o}_k = [&\boldsymbol{\psi}_k, \phi_k],~k\in[0,K-1],
\label{eq:state}
\end{aligned}
\end{equation}
where $\mathcal{X}$ represents the entire state\textcolor{edit}{,} and $\mathbf{x}_i$ represents the body state in the $i$-th sliding window, which is made up of the following parameters: position, quaternion, velocity, and biases of the accelerometer and gyroscope. In addition, the entire state includes the inverse depths of point features, which are represented as $\lambda_j,~j\in[0,J-1]$. In this paper, lines expressed in the orthonormal representations are newly added as $\mathbf{o}$. $I$, $J$, and $K$ are the numbers of sliding window, point features, and line features, respectively.

\begin{figure}[t]
    \centering
    \includegraphics[width=1\linewidth]{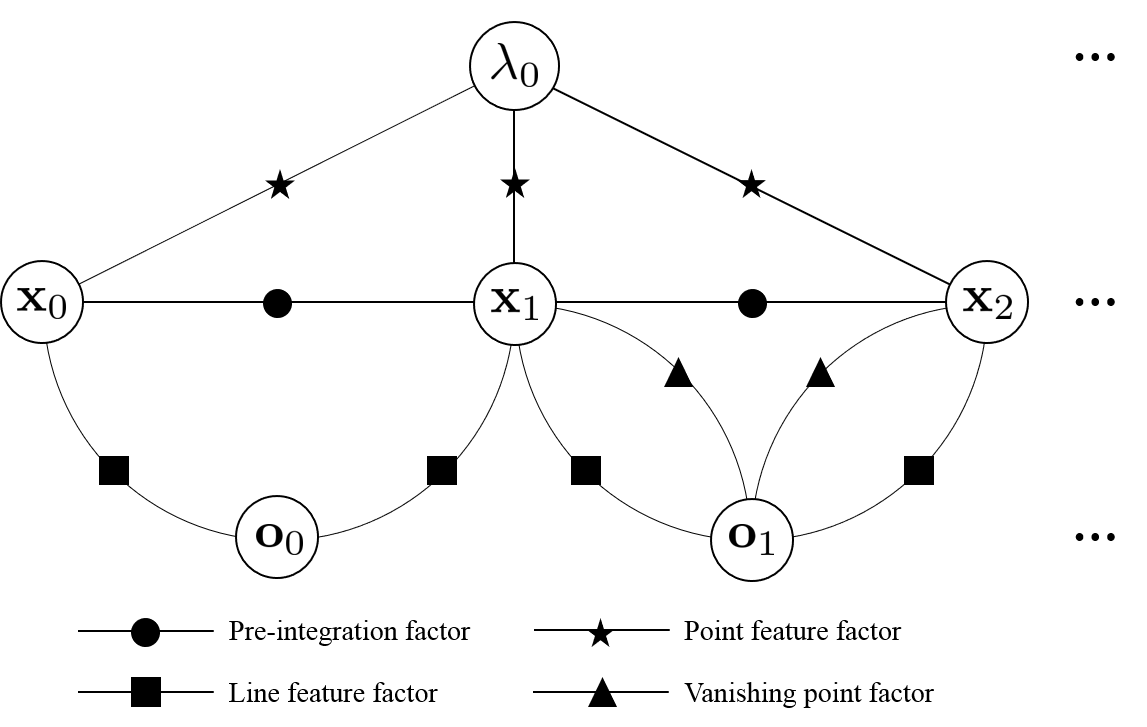}
    \caption{An example of \textcolor{edit}{a} factor graph for UV-SLAM. For $\mathbf{o}_0$, only the line feature factor is used as a \textcolor{edit}{nonstructural line}. Therefore, only line feature factor is employed. On the other hand, $\mathbf{o}_1$ is a line with structural regularity and both line feature factor and the vanishing point factor are used.}
    \label{fig:factor_graph}
    \vspace{-5mm}
\end{figure}

\subsection{UV-SLAM}
Employing defined states in \eqref{eq:state}, the entire objective for optimization is as follows:
\begin{equation}
\resizebox{.9\hsize}{!}{$
\begin{gathered}
\min_{\mathcal{X}} \Bigg\{
\parallel \mathbf{r}_{0} - \mathbf{J}_{0}\mathcal{X} \parallel ^2
\Bigg. \\  \left. 
+ \sum_{i \in \mathcal{B}}\parallel{\mathbf{r}_{I}({\mathbf{z}}^{b_i}_{b_{i+1}}, \mathcal{X})\parallel}_{\Sigma^{b_i}_{b_{i+1}}}^2 
+ \sum_{(i, j) \in \mathcal{P}}\textcolor{black}{\rho_p}\parallel{\mathbf{r}_{p}({\mathbf{z}}^{c_i}_{p_j}, \mathcal{X})\parallel}_{\Sigma^{c_i}_{p_j}}^2
\right. \\ \Bigg. 
+ \sum_{(i, k) \in \mathcal{L}}\textcolor{black}{\rho_l}\parallel{\mathbf{r}_{l}({\mathbf{z}}^{c_i}_{l_k}, \mathcal{X})\parallel}_{\Sigma^{c_i}_{l_k}}^2
+ \sum_{(i, k) \in \mathcal{V}}\textcolor{black}{\rho_v}\parallel{\mathbf{r}_{v}({\mathbf{z}}^{c_i}_{v_k}, \mathcal{X})\parallel}_{\Sigma^{c_i}_{v_k}}^2
\Bigg\},
\label{eq:cost function}
\end{gathered}$}
\end{equation}
where $\mathbf{r}_0$, $\mathbf{r}_I$, $\mathbf{r}_p$, \textcolor{black}{$\mathbf{r}_l$, and $\mathbf{r}_v$} represent marginalization, IMU, point, \textcolor{black}{line, and vanishing point} measurement residuals, respectively. \textcolor{edit}{In addition,} ${\mathbf{z}}^{b_i}_{b_{i+1}}$, ${\mathbf{z}}^{c_i}_{p_j}$, ${\mathbf{z}}^{c_i}_{l_k}$, and ${\mathbf{z}}^{c_i}_{v_k}$ stand for observations of IMU, point, line, and vanishing point, respectively\textcolor{edit}{;} $\mathcal{B}$ is the set of all pre-integrated IMU measurements in a sliding window\textcolor{edit}{;} $\mathcal{P}$, $\mathcal{L}$ and $\mathcal{V}$ are the sets of point, line, and vanishing point measurements in observed frames\textcolor{edit}{;} and $\Sigma^{b_i}_{b_{i+1}}$, $\Sigma^{c_i}_{p_j}$, $\Sigma^{c_i}_{l_k}$ and $\Sigma^{c_i}_{v_k}$ represent IMU, point, line, and vanishing point measurement covariance matrices, respectively. \textcolor{black}{$\rho_p$, $\rho_l$, and $\rho_v$ mean loss functions of the point, line, and vanishing point measurements, respectively. $\rho_p$ and $\rho_l$ are set to the Huber norm function \cite{huber1992robust} and $\rho_v$ is set to the inverse tangent function because of the vanishing point measurement model's unbound problem.} An example of the factor graph for the defined cost function is shown in Fig.~\ref{fig:factor_graph}. If there is no vanishing point measurement for a specific line, only the line feature factor is used as in the case of $\mathbf{o}_0$. If a specific line has corresponding vanishing point measurement, the line feature and vanishing point factors are employed as in the case of $\mathbf{o}_1$. For the optimization process, Ceres Solver\cite{ceres-solver} is used.

\subsection{Line Measurement Model}
First, the re-projection of the 3D line $\mathbf{L}$ in the Pl\"{u}cker coordinate is as follows:
\begin{equation}
\begin{aligned}
\mathbf{l}^c 
&= \begin{bmatrix}l_1 \\ l_2 \\ l_3\end{bmatrix}
= \mathbf{K}'{\mathbf{n}}^c
= f_xf_y(\mathbf{K}^{-1})^\top{\mathbf{n}}^c \\
&= \begin{bmatrix}
f_y & 0 & 0 \\
0 & f_x & 0 \\
-f_yc_x & -f_xc_y & f_xf_y
\end{bmatrix}
{\mathbf{n}}^c = \mathbf{n}^c,
\label{eq:line_projection}
\end{aligned}
\end{equation}
where ${\mathbf{l}}$, $\mathbf{K}'$ and $\mathbf{K}$ represent the re-projected line, the projection matrix of a line feature, and the camera's intrinsic parameter, respectively. ($f_x$, $f_y$) and ($c_x$, $c_y$) denote image's focal lengths and principal points, respectively. Because the proposed algorithm applies \textcolor{edit}{to} a normalized plane, $\mathbf{K}$ and $\mathbf{K}'$ are identity matrices. As a result, the re-projected line is equal to the normal vector in the proposed method.

\begin{figure}[t]
    \centering
    \includegraphics[width=1\linewidth]{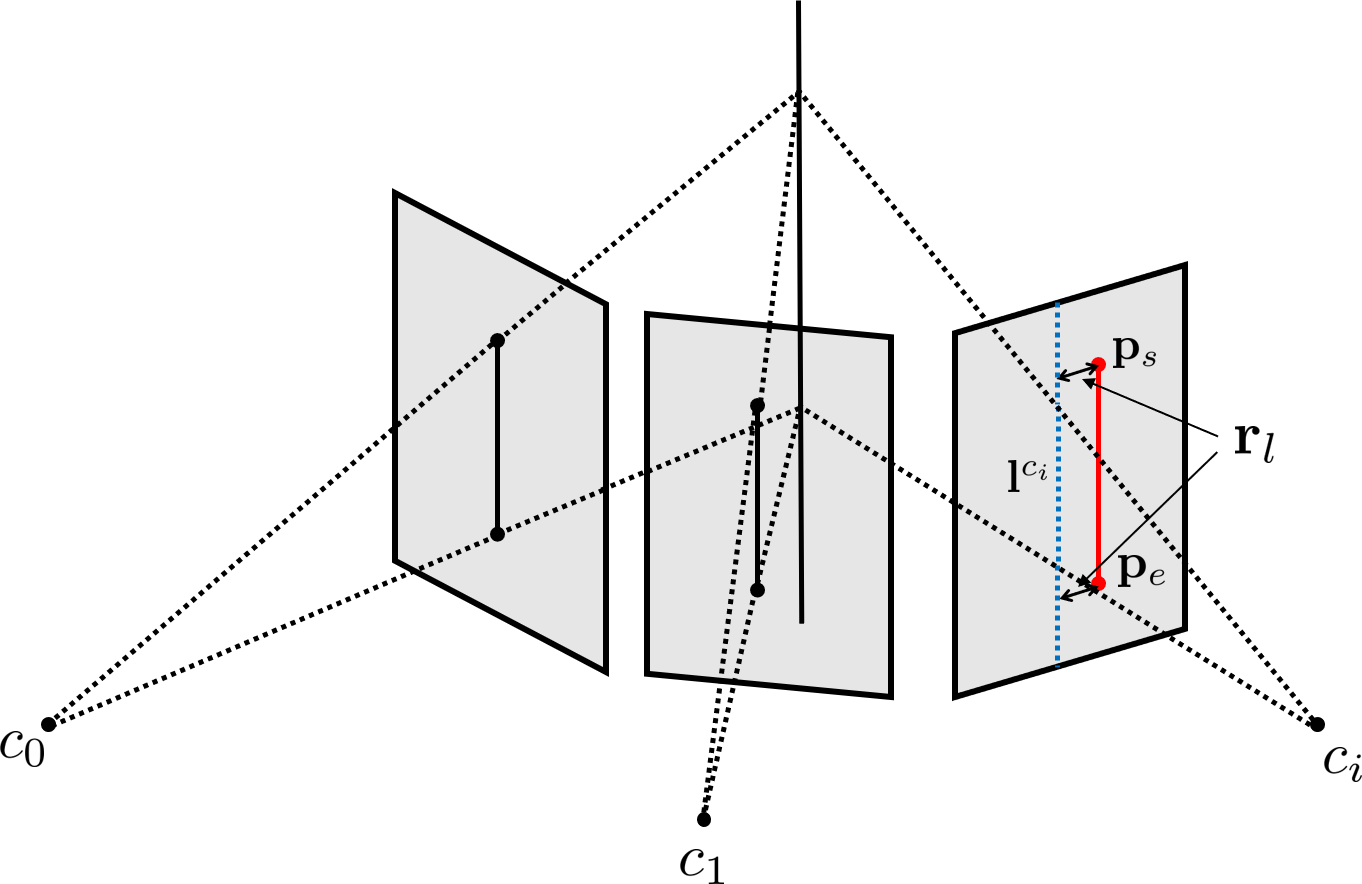}
    \caption{Illustration of line residual. The red solid and blue dashed lines on the image represent observation and re-projected estimation, respectively. The 3D line obtained from triangulation is re-projected into a new frame. \textcolor{edit}{Afterward}, the distance between both endpoints of the observed line and the re-projected line is defined as the residual $\mathbf{r}_l$ of the line.}
    \label{fig:line_residual}
    \vspace{-5mm}
\end{figure}

As shown in Fig.~\ref{fig:line_residual}, \textcolor{edit}{t}he residual of the line measurement model is defined as the following re-projection error:
\begin{equation}
\begin{gathered}
\mathbf{r}_{l} = 
\begin{bmatrix}
d(\mathbf{p}_s, \mathbf{l}^c) \\ d(\mathbf{p}_e, \mathbf{l}^c)
\end{bmatrix}, 
\end{gathered}
\end{equation}
where
\begin{equation}
\begin{gathered}
d(\mathbf{p}, \mathbf{l}^c) = \cfrac{\mathbf{p}^{\top}\mathbf{l}^c}{l_d},~l_d = \sqrt{l_1^2 + l_2^2}, \\
\mathbf{p}_s = (u_s, v_s, 1),~\mathbf{p}_e = (u_e, v_e, 1), \\
\end{gathered}
\end{equation}
and $\mathbf{r}_{l}$ denotes the line residual and $d$ denotes the distance between both endpoints of the observed line and the re-projected line. $\mathbf{p}_s$ and $\mathbf{p}_e$ are the endpoints of the observed line in the image. The corresponding Jacobian matrix with respect to the 3D line can be represented by the body state change, $\delta\mathbf{x}$, and the orthonormal representation change, $\delta\mathbf{o}$, as follows:   
\begin{equation}
\mathbf{J}_{l}=
\cfrac{\partial\mathbf{r}_{l}}{\partial\mathbf{l}^{c}} 
\cfrac{\partial\mathbf{l}^{c}}{\partial\mathbf{L}^{c}} 
\begin{bmatrix}
\cfrac{\partial\mathbf{L}^{c}}{\partial\delta\mathbf{\mathbf{x}}} &
\cfrac{\partial\mathbf{L}^{c}}{\partial\mathbf{L}^{w}} 
\cfrac{\partial\mathbf{L}^{w}}{\partial\delta\mathbf{\mathbf{o}}}
\end{bmatrix}\textcolor{edit}{,}
\label{eq:line_jacobian}
\end{equation}
with 
\begin{equation}
\begin{aligned}
&\cfrac{\partial\mathbf{r}_{l}}{\partial\mathbf{l}^{c}} = 
\begin{bmatrix}
\cfrac{-l_1(\mathbf{p}^{\top}_s\mathbf{l}^c)}{l_d^3}+\cfrac{u_s}{l_d} & \cfrac{-l_2(\mathbf{p}^{\top}_s\mathbf{l}^c)}{l_d^3}+\cfrac{v_s}{l_d} &
\cfrac{1}{l_d} \\
\cfrac{-l_1(\mathbf{p}^{\top}_e\mathbf{l}^c)}{l_d^3}+\cfrac{u_e}{l_d} & \cfrac{-l_2(\mathbf{p}^{\top}_e\mathbf{l}^c)}{l_d^3}+\cfrac{v_e}{l_d} &
\cfrac{1}{l_d}
\end{bmatrix}_{2\times3}, \\
&\cfrac{\partial\mathbf{l}^{c}}{\partial\mathbf{L}^{c}} = 
\begin{bmatrix}
\mathbf{K'} & \mathbf{0}_{3\times3}
\end{bmatrix}_{3\times6}, \\
&\cfrac{\partial\mathbf{L}^{c}}{\partial\delta\mathbf{\mathbf{x}}} = \\
&(\mathcal{T}^b_c)^{-1}
\left[
\begin{matrix}
\begin{matrix}
(\mathbf{R}^w_b)^{\top}[\mathbf{d}^w]_\times \\
\mathbf{0}_{3\times3}
\end{matrix} &
\begin{matrix}
[(\mathbf{R}^w_b)^{\top}(\mathbf{n}^w + [\mathbf{d}^w]_{\times}\mathbf{p}^w_b)]_\times \\
\mathbf{0}_{3\times3}
\end{matrix}
\end{matrix} 
\right. \\ 
&\qquad\qquad\qquad\qquad\qquad
\left.
\begin{matrix}
\begin{matrix}
\mathbf{0}_{6\times3}
\end{matrix} &
\begin{matrix}
\mathbf{0}_{6\times3}
\end{matrix} &
\begin{matrix}
\mathbf{0}_{6\times3}
\end{matrix} &
\end{matrix}
\right]_{6\times15}, \\
&\cfrac{\partial\mathbf{L}^{c}}{\partial\mathbf{L}^{w}} 
\cfrac{\partial\mathbf{L}^{w}}{\partial\delta\mathbf{\mathbf{o}}} = \\
&(\mathcal{T}^b_c)^{-1}
\begin{bmatrix}
\mathbf{0}_{3\times1} & -w_1\mathbf{u}_3 & w_1\mathbf{u}_2 & -w_2\mathbf{u}_1 \\
w_2\mathbf{u}_3 & \mathbf{0}_{3\times1} & -w_2\mathbf{u}_1 & w_1\mathbf{u}_2
\end{bmatrix}_{6\times4}, \\
\end{aligned}
\label{eq:line_jacobian_sub}
\end{equation}
where 
\begin{equation}
\begin{aligned}
&\mathbf{U} = 
\begin{bmatrix}
\mathbf{u}_1 & \mathbf{u}_2 & \mathbf{u}_3
\end{bmatrix} =
\begin{bmatrix}
\cfrac{\mathbf{n}}{\parallel{\mathbf{n}}\parallel} & 
\cfrac{\mathbf{d}}{\parallel{\mathbf{d}}\parallel} &
\cfrac{\mathbf{n}\times\mathbf{d}}{\parallel{\mathbf{n}\times\mathbf{d}}\parallel}
\end{bmatrix}, \\
&\mathbf{w} = 
\begin{bmatrix}
w_1 \\ w_2
\end{bmatrix} =
\cfrac{1}{\sqrt{\parallel{\mathbf{n}}\parallel^2 + \parallel{\mathbf{d}}\parallel^2}}
\begin{bmatrix}
\parallel{\mathbf{n}}\parallel \\ \parallel{\mathbf{d}}\parallel
\end{bmatrix},
\end{aligned}
\end{equation}
and $\mathcal{T}^b_c$ is a transformation matrix from the camera coordinate to the body coordinate in the Pl\"{u}cker coordinate.

\subsection{Vanishing Point Measurement Model}
\begin{figure}[t]
    \centering
    \includegraphics[width=1\linewidth]{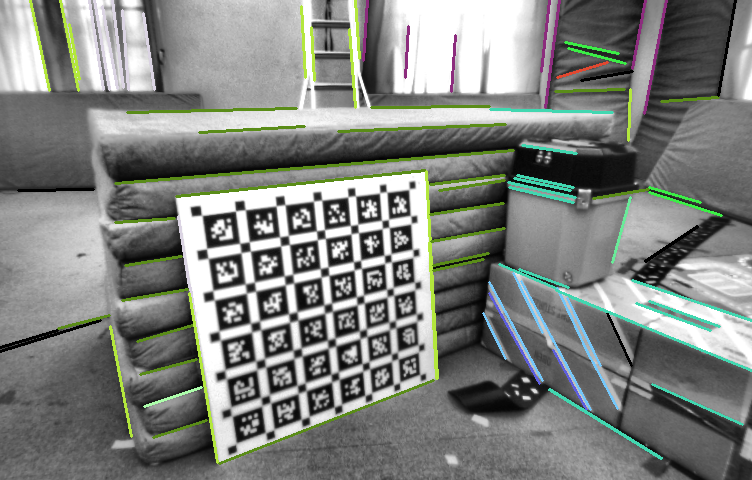}
    \caption{Image of clustered lines using vanishing points. The lines with the same vanishing point are expressed in the same color. Usually, three or more vanishing points can be extracted using J-linkage.}
    \label{fig:vp_extraction}
    \vspace{-5mm}
\end{figure}

After the vanishing points are calculated, the observed line features use the corresponding vanishing points as new observations. An example of clustering lines through vanishing points is shown in Fig.~\ref{fig:vp_extraction}. The lines with the same vanishing point are expressed in the same color. 

\begin{figure}[t]
    \centering
    \includegraphics[width=1\linewidth]{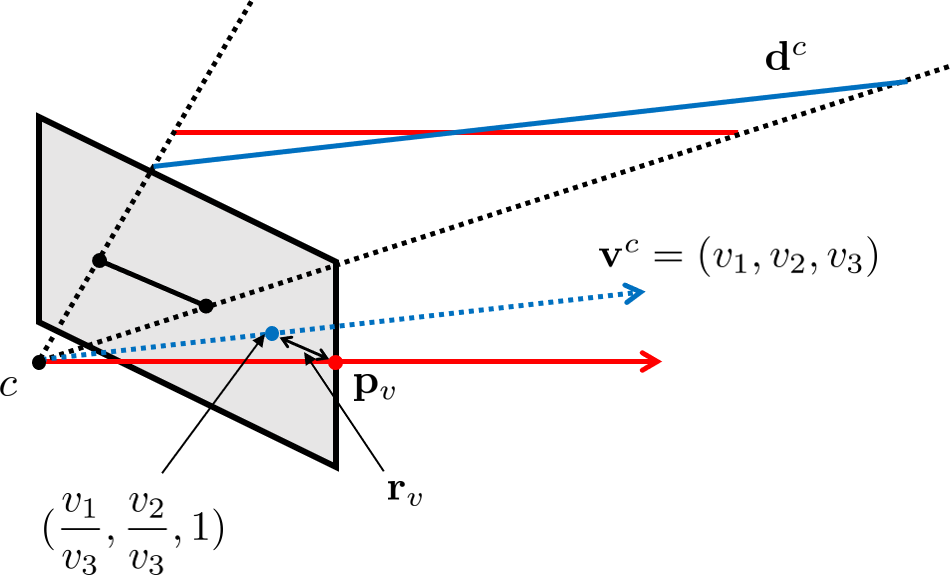}
    \caption{Illustration of the vanishing point residual. The points where the red solid and blue dashed lines intersect with the image represent observation and estimation\textcolor{edit}{, respectively}. In vanishing point estimation, the direction vector of the line is used. The vanishing point difference between the observation from the observed line and the estimation from the 3D line is defined as the residual.}
    \label{fig:vp_residual}
    \vspace{-5mm}
\end{figure}

To estimate the vanishing points, the point on the 3D line is expressed in a homogeneous coordinate as follows\cite{hartley2003multiple}:
\begin{equation}
\mathbf{V}(t) = \mathbf{V}_0 + t\mathbf{D} = 
\begin{bmatrix}
x_0 + td_1 \\ y_0 + td_2 \\ z_0 + td_3 \\ 1
\end{bmatrix}, t \in (0,\infty),
\end{equation}
where
\begin{equation}
\begin{gathered}
\mathbf{V}_0 = 
\begin{bmatrix}
x_0 & y_0 & z_0 & 1
\end{bmatrix}^\top, \\
\mathbf{D} = 
\begin{bmatrix}
{\mathbf{d}^c}^\top, 0
\end{bmatrix}^\top = 
\begin{bmatrix}
d_1, d_2, d_3, 0
\end{bmatrix}^\top,
\end{gathered}
\end{equation}
and $\mathbf{V}_0$ represents a point on the 3D line. Then, the vanishing point equals the projection of a point at infinity on the 3D line as follows:
\begin{equation}
\begin{gathered}    
\mathbf{v}^c = 
\begin{bmatrix}
v_1 \\ v_2 \\ v_3
\end{bmatrix}  
= \lim_{t \to \infty}\mathbf{P}(\mathbf{V}_0 + t\mathbf{D})
= \mathbf{K}\mathbf{d}^c = \mathbf{d}^c,
\end{gathered}
\end{equation}
where $
\begin{gathered}
\mathbf{P} = \mathbf{K}
\left[\begin{array}{@{}c|c@{}}
\mathbf{I} & \mathbf{0}
\end{array}\right] \\
\end{gathered}$ is a camera projection matrix. In the UV-SLAM, the vanishing point from the line is equal to the direction vector of the line. The vanishing point estimation is calculated by the intersection of the $\mathbf{v}^{c}$ and the image plane, as shown in Fig.~\ref{fig:vp_residual}. Finally, the vanishing point residual is as follows:
\begin{equation}
\mathbf{r}_{v} = 
\mathbf{p}_v - \cfrac{1}{v_3}
\begin{bmatrix}
v_1 \\ v_2
\end{bmatrix},
\end{equation}
where $\mathbf{r}_{v}$ and $\mathbf{p}_v$ represent the vanishing point residual and the vanishing point observation, respectively. The corresponding Jacobian matrix with respect to the vanishing point can be obtained in terms of $\delta\mathbf{x}$ and $\delta\mathbf{o}$ as follows:
\begin{equation}
\mathbf{J}_{v}=
\cfrac{\partial\mathbf{r}_{v}}{\partial\mathbf{v}^{c}} 
\cfrac{\partial\mathbf{v}^{c}}{\partial\mathbf{L}^{c}} 
\begin{bmatrix}
\cfrac{\partial\mathbf{L}^{c}}{\partial\delta\mathbf{\mathbf{x}}} &
\cfrac{\partial\mathbf{L}^{c}}{\partial\mathbf{L}^{w}} 
\cfrac{\partial\mathbf{L}^{w}}{\partial\delta\mathbf{\mathbf{o}}}
\end{bmatrix}
\label{eq:vp_jacobian}
\end{equation}
where
\begin{equation}
\begin{aligned}
&\cfrac{\partial\mathbf{r}_{v}}{\partial\mathbf{v}^{c}} = 
\begin{bmatrix}
-\cfrac{1}{v_3} & 0 & \cfrac{v_1}{v_3^2} \\
0 & -\cfrac{1}{v_3} & \cfrac{v_2}{v_3^2}
\end{bmatrix}_{2\times3}, \\
&\cfrac{\partial\mathbf{v}^{c}}{\partial\mathbf{L}^{c}} = 
\begin{bmatrix}
\mathbf{0}_{3\times3} & \mathbf{K} 
\end{bmatrix}_{3\times6}.
\end{aligned}
\label{eq:vp_jacobian_sub}
\end{equation}

\section{Fisher Information Matrix Rank Analysis}\label{sec:FIM rank analysis}
We rigorously analyze the observability of line features through Fisher information matrix (FIM) rank analysis. If the FIM is singular, the system is unobservable\cite{bar2004estimation}. Wang \textit{et al.} proved that the Jacobian matrix used in the FIM calculation must satisfy the full column rank condition for the FIM to satisfy nonsingularity\cite{wang2008observability}. We also use this approach to analyze the observability of the proposed method.

First, the FIM of the line measurement in the orthonormal representation is as follows:
\begin{equation}
\begin{gathered}
\mathbf{H}_{l}^{\delta\mathbf{o}} =
\mathbf{J}_{l}^{\delta\mathbf{o}\top}
\boldsymbol{\Omega}_{l}^{\delta\mathbf{o}}
\mathbf{J}_{l}^{\delta\mathbf{o}},
\end{gathered}
\end{equation}
where
\begin{equation}
\begin{aligned}
\mathbf{J}_{l}^{\delta\mathbf{o}}= &
\cfrac{\partial\mathbf{r}_{l}}{\partial\mathbf{l}^{c}} 
\cfrac{\partial\mathbf{l}^{c}}{\partial\mathbf{L}^{c}} 
\cfrac{\partial\mathbf{L}^{c}}{\partial\mathbf{L}^{w}} 
\cfrac{\partial\mathbf{L}^{w}}{\partial\delta\mathbf{\mathbf{o}}} \\
=& \begin{bmatrix}
0 & j_{l_{12}} & j_{l_{13}} & j_{l_{14}} \\
0 & j_{l_{22}} & j_{l_{23}} & j_{l_{24}}
\end{bmatrix},
\end{aligned}
\label{eq:line_jacobian_rank}
\end{equation}
and $j_{l_{pq}}$ is the non-zero element in the $p$-th row and the $q$-th column of $\mathbf{J}_{l}^{\delta\mathbf{o}}$, which is obtained by substituting \eqref{eq:line_jacobian_sub} into \eqref{eq:line_jacobian_rank}. $\boldsymbol{\Omega}_{l}^{\delta\mathbf{o}}$ represents the inverse of covariance matrix of the line observation. \textcolor{edit}{Because} $\frac{\partial\mathbf{l}^{c}}{\partial\mathbf{L}^{c}}$, $\frac{\partial\mathbf{L}^{c}}{\partial\mathbf{L}^{w}}$, and $\frac{\partial\mathbf{L}^{w}}{\partial\delta\mathbf{\mathbf{o}}}$ are full rank matrices from \eqref{eq:line_jacobian_sub}, \textcolor{edit}{the} rank of $\mathbf{J}_{l}^{\delta\mathbf{o}}$ is determined by $\frac{\partial\mathbf{r}_{l}}{\partial\mathbf{l}^{c}}$ in \eqref{eq:line_jacobian_rank}. From \eqref{eq:line_jacobian_sub}, the maximum rank of $\frac{\partial\mathbf{r}_{l}}{\partial\mathbf{l}^{c}}$ is 2, and the case \textcolor{edit}{in which} the rank becomes 1 is as follows:
\begin{equation}
\cfrac{l_2}{l_1} = \cfrac{v_s-v_e}{u_s-u_e}.
\label{eq:exceptional_case}
\end{equation}
However, \textcolor{edit}{because} a line in the orthonormal representation has four parameters, the observability of the line cannot be guaranteed with the line measurement model alone. To solve this problem, a new observation on the line other than both endpoints is introduced as follows:
\begin{equation}
\mathbf{p}_l = \alpha\mathbf{p}_s + (1-\alpha)\mathbf{p}_e, \alpha\in(0,1),
\end{equation}
where $\mathbf{p}_l$ represents the new observation. However, despite adding a new observation, the rank of $\mathbf{J}_{l}^{\delta\mathbf{o}}$ is up to 2. Therefore, the line features using only the line measurement model are still not observable.

From a new perspective, we propose to calculate the FIM with the vanishing point measurements. The FIM of the vanishing point measurement in the orthonormal representation is as follows: 
\begin{equation}
\begin{gathered}
\mathbf{H}_{v}^{\delta\mathbf{o}} =
\mathbf{J}_{v}^{\delta\mathbf{o}\top}
\boldsymbol{\Omega}_{v}^{\delta\mathbf{o}}
\mathbf{J}_{v}^{\delta\mathbf{o}},
\end{gathered}
\end{equation}
where
\begin{equation}
\begin{aligned}
\mathbf{J}_{v}^{\delta\mathbf{o}} =&
\cfrac{\partial\mathbf{r}_{v}}{\partial\mathbf{v}^{c}} 
\cfrac{\partial\mathbf{v}^{c}}{\partial\mathbf{L}^{c}} 
\cfrac{\partial\mathbf{L}^{c}}{\partial\mathbf{L}^{w}} 
\cfrac{\partial\mathbf{L}^{w}}{\partial\delta\mathbf{\mathbf{o}}} \\
=& \begin{bmatrix}
j_{v_{11}} & 0 & j_{v_{13}} & j_{v_{14}} \\
j_{v_{21}} & 0 & j_{v_{23}} & j_{v_{24}}
\end{bmatrix},
\end{aligned}
\label{eq:vp_jacobian_rank}
\end{equation}
and $j_{v_{pq}}$ is the non-zero element in the $p$-th row and the $q$-th column of $\mathbf{J}_{v}^{\delta\mathbf{o}}$, which is obtained by substituting \eqref{eq:vp_jacobian_sub} into \eqref{eq:vp_jacobian_rank}. $\boldsymbol{\Omega}_{v}^{\delta\mathbf{o}}$ represents the inverse of covariance matrix of the vanishing point observation. Similarly, all matrices are full rank except $\frac{\partial\mathbf{r}_{v}}{\partial\mathbf{v}^{c}}$. Therefore, the rank of $\mathbf{J}_{v}^{\delta\mathbf{o}}$ is 2, which can be obtained from the rank of $\frac{\partial\mathbf{r}_{v}}{\partial\mathbf{v}^{c}}$ in \eqref{eq:vp_jacobian_sub}. 

By eigenvalue decomposition, the FIM considering both the line measurement and the vanishing point measurement can be obtained as follows:
\begin{equation}
\begin{aligned}
\mathbf{H}^{\delta\mathbf{o}} 
&= \mathbf{H}_{l}^{\delta\mathbf{o}} + \mathbf{H}_{v}^{\delta\mathbf{o}} \\
&= \mathbf{J}_{l}^{\delta\mathbf{o}\top}\boldsymbol{\Omega}_{l}^{\delta\mathbf{o}}\mathbf{J}_{l}^{\delta\mathbf{o}} + \mathbf{J}_{v}^{\delta\mathbf{o}\top}\boldsymbol{\Omega}_{v}^{\delta\mathbf{o}}\mathbf{J}_{v}^{\delta\mathbf{o}} \\
&= \mathbf{J}^{\delta\mathbf{o}\top}\boldsymbol{\Omega}^{\delta\mathbf{o}}\mathbf{J}^{\delta\mathbf{o}},
\end{aligned}
\end{equation}
where 
\begin{equation}
\begin{gathered}
\mathbf{J}^{\delta\mathbf{o}} = 
\begin{bmatrix}
0 & j_{l_{12}} & j_{l_{13}} & j_{l_{14}} \\
0 & j_{l_{22}} & j_{l_{23}} & j_{l_{24}} \\ 
j_{v_{11}} & 0 & j_{v_{13}} & j_{v_{14}} \\
j_{v_{21}} & 0 & j_{v_{23}} & j_{v_{24}}
\end{bmatrix},\\
\boldsymbol{\Omega}^{\delta\mathbf{o}} = 
\begin{bmatrix}
\boldsymbol{\Omega}_l^{\delta\mathbf{o}} & 0 \\
0 & \boldsymbol{\Omega}_v^{\delta\mathbf{o}}
\end{bmatrix}.
\end{gathered}
\end{equation}
At this time, the rows of the line measurement and the vanishing point measurement are independent in $\mathbf{J}^{\delta\mathbf{o}}$. Therefore, the rank of $\mathbf{J}^{\delta\mathbf{o}}$ is 4\textcolor{edit}{,} except for the case of \eqref{eq:exceptional_case}. We can confirm that the line features become fully observable by additionally using the vanishing point meausrement model.

\section{Experimental Results}\label{sec:experimental results}
The experiments were carried out on an Intel Core i7-9700K processor with 32GB of RAM. Using the EuRoC micro aerial vehicle (MAV) datasets \cite{burri2016euroc}, we tested the state-of-the-art algortihms and the UV-SLAM. Each dataset offers a varied level of complexity depending on factors like lighting, texture, and MAV speed. Therefore, the datasets were appropriate to validate the performance of the proposed method. 

\begin{figure*}[ht!]
    \centering
    \subfigure[]{\label{fig:1}\includegraphics[width=0.49\linewidth]{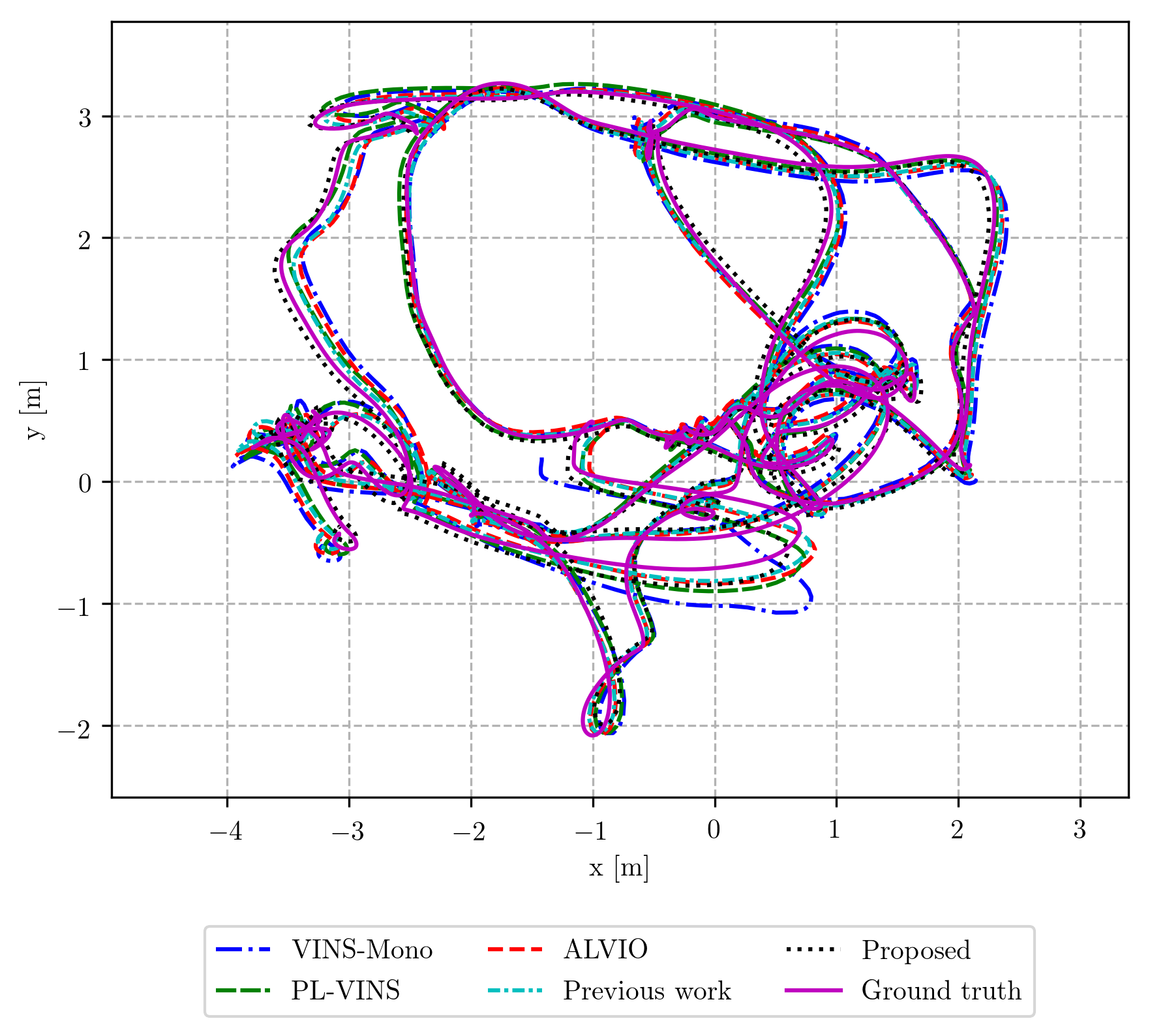}}
    \subfigure[]{\label{fig:2}\includegraphics[width=0.49\linewidth]{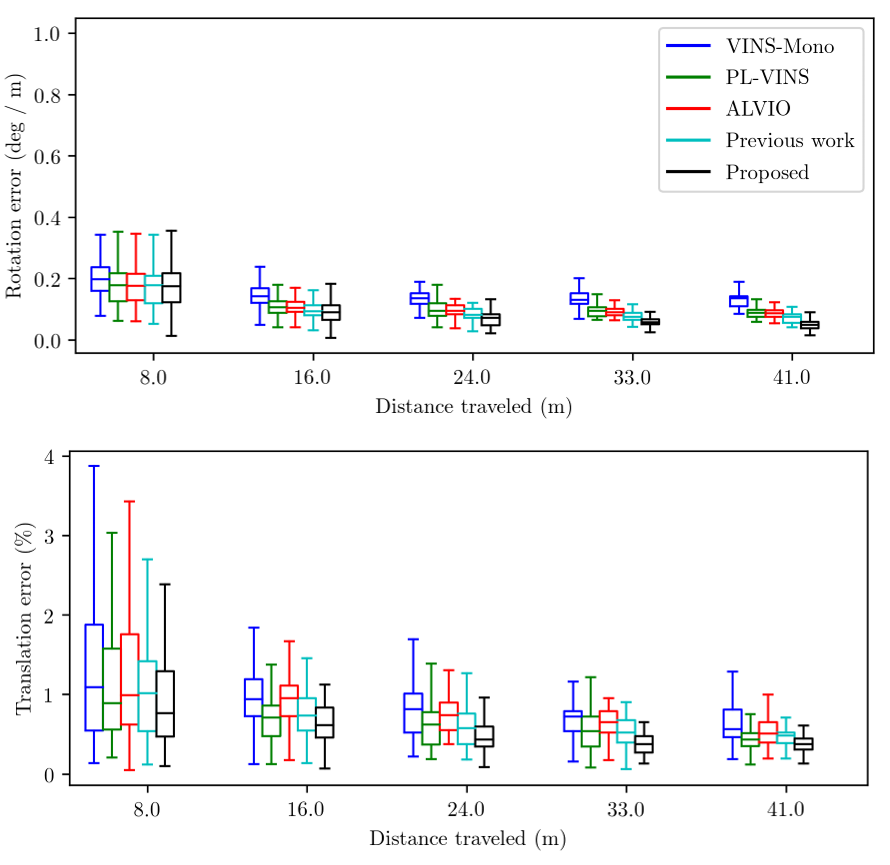}}
    \caption{The top views of \subref{fig:1} trajectory and \subref{fig:2} boxplot of RMSE according to distance traveled for VINS-Mono, PL-VINS, ALVIO, \textcolor{black}{our previous work\cite{lim2021avoiding}}, and UV-SLAM for {\tt{V2\_02\_medium}} in the EuRoC datasets.}
    \label{fig:rpg_result}
\end{figure*}

We compared the localization accuracy of the proposed method with that of VINS-Mono which is our base algorithm. In addition, we also compared PL-VINS\cite{fu2020pl}, ALVIO, \textcolor{black}{and our previous work} which use line features on top of VINS-Mono. The parameters of compared algorithms are set to the default values in the open-source codes. We employed the rpg trajectory evaluation tool\cite{zhang2018tutorial}. Table~\ref{table:error} shows the translational root mean square error (RMSE) for the EuRoC datasets. The proposed method has better performance than state-of-the-art algorithms in all datasets. In particular, the proposed algorithm shows 32.3\%, 23.8\%, 26.4\%, and \textcolor{black}{17.6\%} smaller average error than VINS-Mono, PL-VINS, ALVIO, \textcolor{black}{and our previous work}, respectively. More accurate results could be obtained in the proposed algorithm because the line features become fully observable using the vanishing point measurements. The results for \textcolor{edit}{trajectory, rotation error, and translation error of} {\tt{V2\_02\_medium}} in \textcolor{edit}{the} EuRoC datasets \textcolor{edit}{are} shown in Fig.~\ref{fig:rpg_result}. 

\begin{table}[t]
\centering
\caption{Translational RMSE without loop closing for the EuRoC Datasets (Unit: m)}
\label{table:error}
\resizebox{\linewidth}{!}{
\begin{tabular}{*{6}{c}}\toprule
Translation & \multirow{2}{*}{VINS-Mono} & \multirow{2}{*}{PL-VINS} & \multirow{2}{*}{ALVIO} & \textcolor{black}{Our method} & \multirow{2}{*}{UV-SLAM} \\
RMSE &  &  &  & \textcolor{black}{in \cite{lim2021avoiding}} &  \\ \midrule
\tt{MH\_01\_easy}      & 0.159 & 0.164 & 0.148 & 0.142 & \textbf{0.139}  \\
\tt{MH\_02\_easy}      & 0.140 & 0.174 & 0.136 & 0.126 & \textbf{0.094}  \\
\tt{MH\_03\_medium}    & 0.225 & \textbf{0.187} & 0.209 & 0.198 & 0.189  \\
\tt{MH\_04\_difficult} & 0.408 & 0.335 & 0.389 & 0.301 & \textbf{0.261}  \\
\tt{MH\_05\_difficult} & 0.312 & 0.347 & 0.317 & 0.293 & \textbf{0.188}  \\
\tt{V1\_01\_easy}      & 0.094 & 0.071 & 0.085 & 0.087 & \textbf{0.067}  \\
\tt{V1\_02\_medium}    & 0.115 & 0.086 & 0.075 & 0.072 & \textbf{0.070}  \\
\tt{V1\_03\_difficult} & 0.203 & 0.152 & 0.200 & 0.156 & \textbf{0.109}  \\
\tt{V2\_01\_easy}      & 0.099 & 0.090 & 0.094 & 0.098 & \textbf{0.085}  \\
\tt{V2\_02\_medium}    & 0.161 & 0.120 & 0.133 & \textbf{0.103} & 0.112  \\
\tt{V2\_03\_difficult} & 0.341 & 0.278 & 0.288 & 0.277 & \textbf{0.213}  \\
\bottomrule 
\end{tabular}}
\end{table}

\begin{figure*}[h!]
    \centering
    \subfigure[]{\label{fig:mapping_result:a}\includegraphics[width=0.32\linewidth]{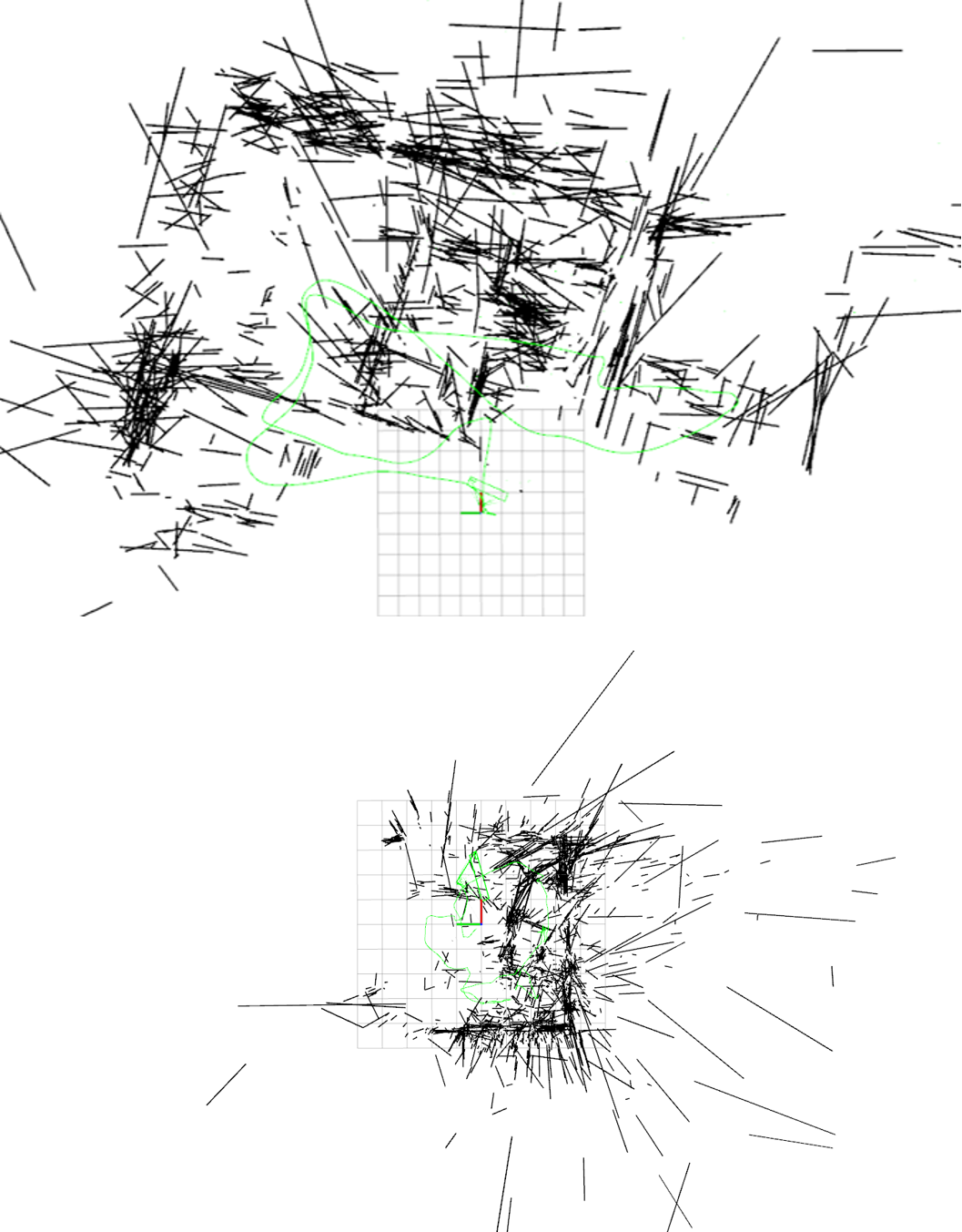}}
    \subfigure[]{\label{fig:mapping_result:b}\includegraphics[width=0.32\linewidth]{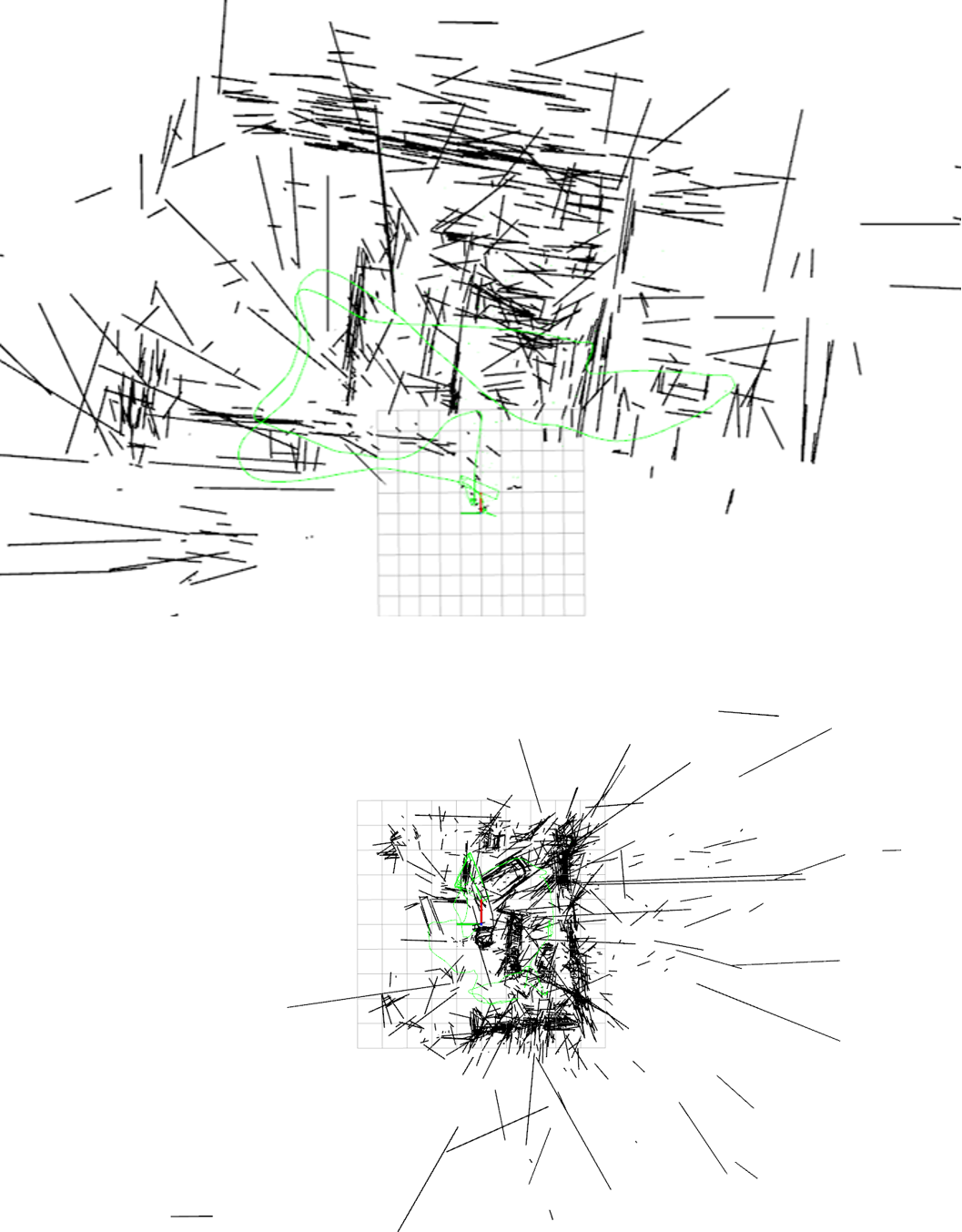}}
    \subfigure[]{\label{fig:mapping_result:c}\includegraphics[width=0.32\linewidth]{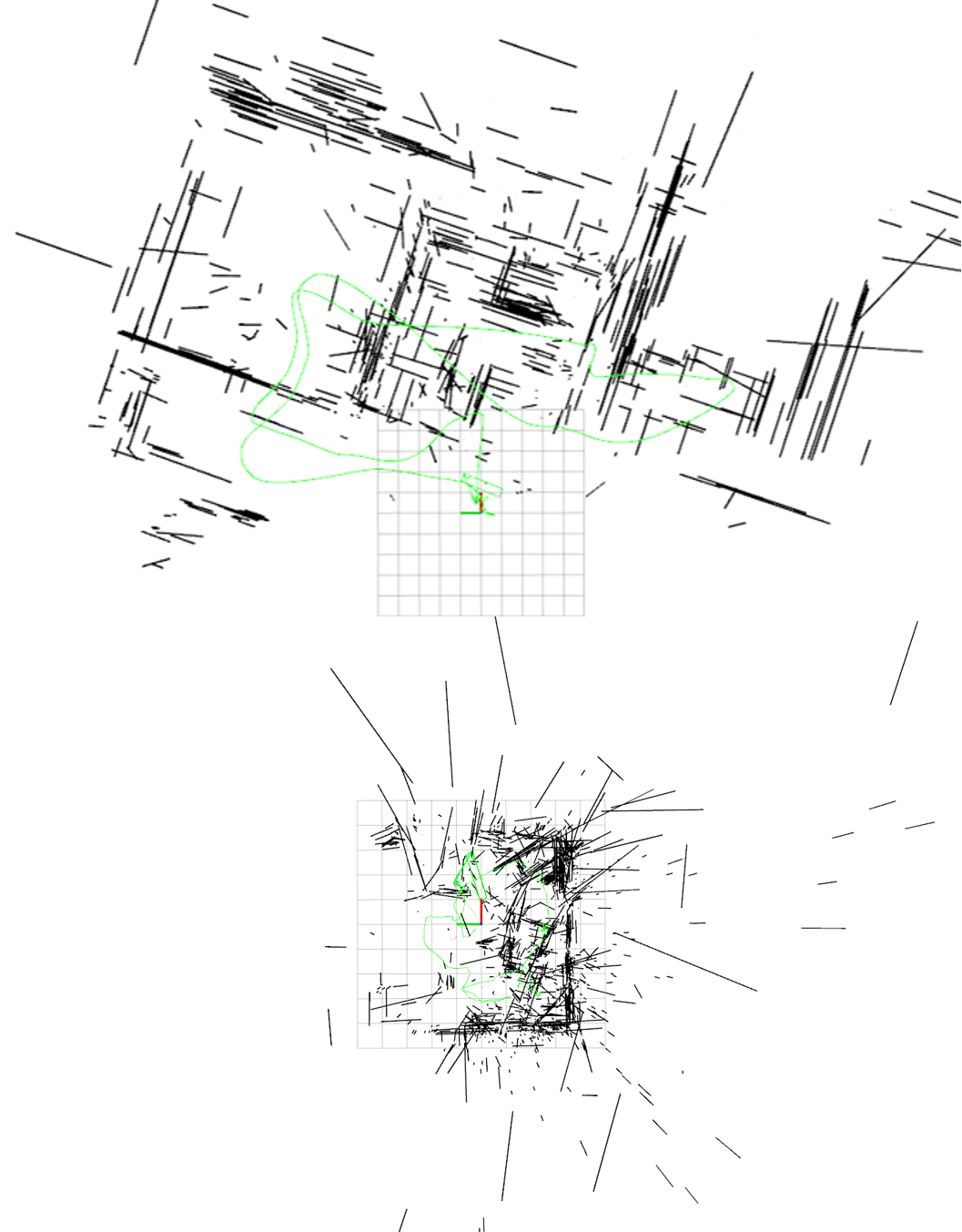}}
    \caption{The top views of line mapping results of \subref{fig:mapping_result:a} ALVIO, \textcolor{black}{\subref{fig:mapping_result:b} our previous work \cite{lim2021avoiding},} and \subref{fig:mapping_result:c} UV-SLAM for {\tt{MH\_05\_difficult}} (top) and {\tt{V2\_01\_easy}} (bottom) in the EuRoC datasets.}
    \label{fig:mapping_result}
\end{figure*}

In addition, \textcolor{edit}{the} mapping results for {\tt{MH\_05\_difficult}} and {\tt{V2\_01\_easy}} in \textcolor{edit}{the} EuRoC datasets are shown in Fig.~\ref{fig:mapping_result}. In the case of ALVIO, the quality of mapping is low due to degenerate lines. \textcolor{black}{In our previous work, degenerate lines were corrected only in pure translational camera motion.} Noteworthily, lines' direction vectors are aligned thanks to the vanishing point measurements in UV-SLAM. \textcolor{black}{All top views of the line mapping results for the EuRoC datasets are available at: \url{https://github.com/url-kaist/UV-SLAM/blob/main/mapping_result.pdf}.}

\textcolor{black}{The average runtime is about 53.528ms for the frontend and 47.086ms for the backend for the EuRoC datasets. UV-SLAM has only about 3ms longer frontend runtime than other algorithms because it extracts vanishing points. Moreover, the runtime of the backend corresponding to optimization is similar to those of other algorithms. This is because the proposed vanishing point measurement model does not use new parameters.}

\section{Conclusion}\label{sec:conclusion}
In summary, we proposed UV-SLAM, which is the unconstrained line-based SLAM using \textcolor{edit}{a} vanishing point measurement. The proposed method can be used without any assumptions such as the Manhattan world. We calculated the residual and Jacobian matri\textcolor{edit}{ces} of \textcolor{edit}{the} vanishing point measurements. Through FIM rank analysis\textcolor{edit}{, we verified} that line's observability is guaranteed by introducing \textcolor{edit}{the} vanishing point measurements into the existing method. In addition, \textcolor{edit}{we showed} that localization accuracy and mapping quality have increased through quantitative and qualitative comparisons with state-of-the-art algorithms. For future work, we will implement mesh or pixel-wise mapping through sparse line mapping from the proposed algorithm.



%




\bibliographystyle{IEEEtran}
\bibliography{reference}
%

%








\end{document}